\pdfoutput=1

\documentclass[11pt]{article}

\usepackage[]{ACL2023}

\usepackage{times}
\usepackage{latexsym}

\usepackage[T1]{fontenc}

\usepackage[utf8]{inputenc}

\usepackage{microtype}

\usepackage{inconsolata}

\usepackage{graphicx}
\usepackage{subcaption}
\usepackage[export]{adjustbox}
\usepackage{rotating}

\usepackage[fleqn]{amsmath}
\usepackage{xcolor}
\usepackage{hyperref}
\usepackage{array}
\usepackage{soul}
\usepackage{algpseudocode}
\usepackage[linesnumbered,ruled,vlined]{algorithm2e}
\SetKwInput{KwInput}{Input}
\SetKwInput{KwOutput}{Output} 
\SetKwInput{KwData}{Data}
\title{Human Inspired Progressive Alignment and Comparative Learning for Grounded Word Acquisition}

\author{Yuwei Bao$^\dagger$ \hspace{30pt} Barrett Martin Lattimer$^\mathsection$\thanks{\ \ Work done during master study at the University of Michigan.}  \hspace{30pt}  Joyce Chai$^\dagger$\\
$^\dagger$Computer Science and Engineering, University of Michigan \hspace{5pt}
$^\mathsection$ASAPP \\
  \texttt{\{yuweibao, lattimer, chaijy\}@umich.edu}  
}

\begin{document}
\maketitle
\begin{abstract}

Human language acquisition is an efficient, supervised, and continual process. In this work, we took inspiration from how human babies acquire their first language, and developed a computational process for word acquisition through comparative learning. Motivated by cognitive findings, we generated a small dataset that enables the computation models to compare the similarities and differences of various attributes, learn to filter out and extract the common information for each shared linguistic label. We frame the acquisition of words as not only the information filtration process, but also as representation-symbol mapping. This procedure does not involve a fixed vocabulary size, nor a discriminative objective, and allows the models to continually learn more concepts efficiently. Our results in controlled experiments have shown the potential of this approach for efficient continual learning of grounded words. 
 
\end{abstract}

\section{Introduction}

Two of the important word acquisition problems are: 1) what must be learned to acquire a word, and 2) how to learn the word?
To the first question, cognitive studies have shown that several critical steps in learning language naturally comes from joint attention establishment \cite{jointAtt}, and symbol grounding \cite{symbolground}. Children's attention are usually redirected through a mother or teacher's guidance, and they learn to map these attended sensory inputs (e.g. color, sound, heat) with their corresponding words or sentences. Living in a rich and diverse world enabled by our multiple body sensors, we learned to filter out the noise and pay attention to specific aspects of an input which we assign linguistic labels to. This attention establishment and information filtration process is the first step of word acquisition. After filtering out the noise, we are left with a mental representation of what a word entails. Just as the word ``car'' triggers certain impressions of a common transportation in head, we store these representations as they could come in handy later when we use them to reason, imagine, and express ourselves. To acquire a word, humans learn to filter out noise to focus on key information from sensory inputs that contributes to its meaning \cite{gentner1983structure, jointAtt}, and store that meaning representation for future use \cite{symbolground, kuehne2000modeling}.

As for the second question, one of the common but under-explored methods is implicit or explicit comparisons. Caretakers may lay out stuffed animals around a baby and name them one by one to differentiate them. In school, teachers may compare different components of the learning material, e.g. ``Today we learn `colors'. This is red/blue/yellow...''. Comparison is the process of finding commonalities and highlighting differences \cite{ref8}. It allows children to attend to matching relational structures of inputs \cite{gentner1983structure}, filter out background noise, and learn to generalize as well as abstract. With comparisons, especially clean well-structured comparisons, children can learn a lot of condensed knowledge efficiently and cultivate their capabilities to tackle noisier challenges outside of the classroom \cite{ref3, anderson2018comparison, ref6}.


From these findings, we propose a new method of word acquisition for artificial intelligent (AI) agents. 
We mimic the classroom learning setting and constructed a small clean dataset named \textbf{SOLA} -- \textbf{S}imulated \textbf{O}bjects for \textbf{L}anguage \textbf{A}cquisition. This dataset allows the model to draw efficient similarity and difference comparisons, learn to filter out noise, pay attention only to key information that contributes to a word meaning, and store these word-representation mappings continually as new words are introduced. 
While a larger scale evaluation is needed in the future, 
through controlled experiments, our preliminary results have demonstrated the potential of this model in efficient continual learning of grounded words.
The dataset and code are available at \url{https://github.com/sled-group/Comparative-Learning}.

The contributions of this work include:
\vspace{-8pt}
\begin{enumerate}
    \item Constructed a small, clean dataset SOLA for studying efficient comparisons.
    \vspace{-5pt}
    \item Framed the acquisition of words as both an information filtration process, and as a representation-symbol mapping.
    \vspace{-5pt}
    \item Proposed a new method of grounded word acquisition through comparative learning
    \vspace{-5pt}
    \item Demonstrated the performance, usability, and generalizability of the acquired representations through multiple tasks.
    \vspace{-10pt}
\end{enumerate}

\begin{figure}[!tbp]
    \includegraphics[width=0.49\textwidth]{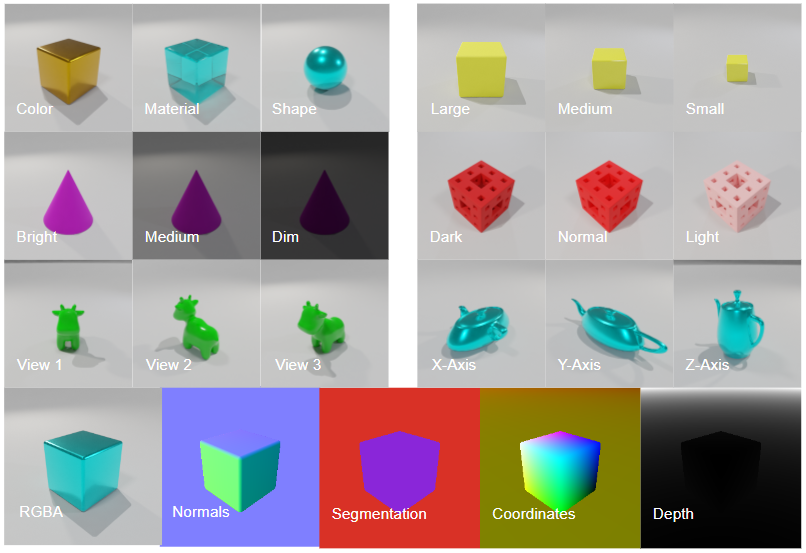}
    \caption{SOLA Dataset}
    \label{sola}
    \vspace{-15pt}
\end{figure}

\section{Related Work}
\subsection{Human Language Acquisition}
Language acquisition is the process of putting linguistic labels onto abstracted features, and structuring them into sentences following publicly recognized grammatical rules to express intention. The simple process of abstracting features, takes input attention filtering and relational generalization to pinpoint the learning concept, and associate them with linguistic labels \cite{symbolground, kuehne2000modeling}. Studies show that the amount of mother-child joint attention facilitation time is positively correlated to a child's early vocabulary size growth \cite{jointAtt}, and that human infants are capable of comparison and abstraction through same/different relation comprehension \cite{anderson2018comparison}. 

Comparison is a central component of human cognition which results in our own uniquely structured knowledge representations \cite{markman1993structural, gentner2017analogical}. The theory of Structure-Mapping predicts that similarity comparison allows subjects to attend to matching relational structures of inputs \cite{gentner1983structure}, highlight the differences, and that human infants are able to learn such relations in very few examples \cite{hespos2020structure}. 

The difficulty of establishing a structural mapping, however, is influenced by the ease of the alignment process \cite{gentner1983structure}. Progressive alignment \cite{hespos2020structure, kotovsky1996comparison} suggests that constructing an alignment among highly similar comparisons can invite young children to reason about relational structures and serve as base knowledge for future hierarchical abstractions and complex characteristic learning.


Our work took inspiration from the above two theories by constructing grouped multimodal samples for similarity and difference comparisons. We also start progressive alignment with highly aligned pairings during early word acquisition.

\vspace{-10pt}
\subsection{Continual Learning}
\vspace{-5pt}
There are two major limitations that current neural network models face. Models either take the large-pretrained approach, throwing as much data as possible during training, and hope to learn everything and achieve AGI \cite{agi} all at once without the need for continual learning. Or models take the architectural \cite{cwr}/ rehearsal \cite{gem}/ replay \cite{gdm}/ regularization \cite{ewc} approaches hoping to retain previously learned knowledge amid newly introduced data distribution shift \cite{ring1998child, nguyen2017variational, schlimmer1986case}. 

Humans, however, are lifelong learners. We are constantly adapting to new environments, learning new concepts \& tasks, and evolving together as a society. Human execution of this process is simple, natural, and cost effective, without catastrophically forgetting previously learned knowledge \cite{van2019three}, nor having to retrain from scratch every time new knowledge is introduced \cite{schlimmer1986case}. 


Our method follows the human learning approach and gradually learns more concepts as they are introduced. We demonstrate the model's resistance against catastrophic forgetting and the data learning efficiency in our experiments.


\subsection{Contrastive Learning}
Contrastive learning is a paradigm that enables models to learn feature representations through contrasting examples without explicit labeling \cite{chen2020simple, wu2018unsupervised, khosla2020supervised}. Contrastive learning uses a single contrastive loss function that pushes similar classes together and dissimilar classes apart \cite{dosovitskiy2014discriminative}. In this paper, we introduce \textbf{Comparative Learning} which adapts the general definition of contrastive learning by \ul{explicitly separating the similarity training from the difference training}. On top of encoding each input as in contrastive learning, we took additional steps to further extract information about similarities and differences separately given the same amount of inputs. Supervised by associated words, we use the similarity batches to learn the process of noise filtration and a shared feature representation. We use the difference batches to refine and differentiate these feature representations. 

\vspace{-5pt}
\subsection{Multimodal Grounding}
A large number of previous works try to draw connections between language and different modalities, such as VILBERT \cite{lu2019vilbert}, LXMERT \cite{tan2019lxmert}, UNITER \cite{chen2020uniter}, OSCAR \cite{li2020oscar}, Vokenization \cite{tan2020vokenization}, and more  \cite{radford2021learning, wang2021simvlm, zhang2021vinvl, cho2021unifying, bao, hill2020grounded, tsimpoukelli2021multimodal}. These models demonstrated their state of the art multimodal representations on a range of downstream tasks, including image/video captioning, image/text retrieval, visual question answering, and text-to-image generation \cite{mdter, zeroc, nscl, multimod4, multimod5}. 
%
A large portion of these works focusd on visual recognition and language production tasks such as captioning, retrieval, and some visual question answering. These works embed visual and textual inputs into the same latent space for similarity comparison and retrieval. These models can learn a great language-vision matching filter, but often do not preserve a grounded concept representation given the linguistic labels.

Another line of works focus on language comprehension and image/video generation. They take a pre-trained language embedding and use it to generate high resolution images, and have achieved extraordinary performance. Notably, \cite{liu2022compositional, du2021unsupervised, du2020compositional} achieved compositional visual generation with energy based models. \cite{brooks2022instructpix2pix} worked on image editing given instructions with paired training images. Also others demonstrated language grounding through compositional text to image generations \cite{feng2022training, pearl2022geocode}. These models rely on great grounded language representations to generate meaningful images.

Our work frames the language acquisition process as both input information filtration and representation learning. A few methods include both parts of this definition. CLIP \cite{radford2021learning} used contrastive learning on massive number of weakly linked image-text pairs to project each modality into the same embedding space, which allows the encoders to filter inputs, and store the representations through text embeddings. Several works including \cite{liu2022compositional, du2021unsupervised, du2020compositional} used a set of energy based models on recognition tasks for input filtration, and iteratively refined the representations through the Langevin dynamics procedure \cite{langevin}. Our work proposes a human inspired approach for word acquisition. We jointly train both the input filtration process and the representations, and map them to their corresponding words through comparative learning. 
\vspace{-5pt}
\section{Dataset} \label{sec_data}
\vspace{-5pt}
Inspired by the classroom teaching setting and the Progressive Alignment theory \cite{kotovsky1996comparison}, we created a new dataset \textbf{SOLA} (\textbf{S}imulated \textbf{O}bjects for \textbf{L}anguage \textbf{A}cquisition). SOLA has little noise and clearly defined attributes to isolate different concepts for efficient sample comparisons and grounded language-feature mapping. We generated SOLA using the open-source simulation software Kubric \cite{greff2021kubric} designed for semi-realistic image/video synthesis.

SOLA (Figure \ref{sola}) contains images of individual simulated objects with three associated learning attributes: color, material, and shape. Each object is a composition of one of 8 colors, 11 shapes, and 4 materials. We also diversify the images by capturing each object at 3 different light settings and 6 different camera angles. A total of 6336 Red Green Blue Alpha (RGBA) images were generated. 
To evaluate the generalizability and robustness of the models on nosier inputs, we also composed a Variation Test set ($D_{test\_v}$) of 989 RGBA images by applying a stretch, shade change, or size transformation. An object in this test set is either stretched along one of the x, y, and z axis, colored with a darker or lighter shade, or shrunk to a medium or small size. Although not used in this work, we rendered the Depth, Surface Normal, Segmentation Map, and Object Coordinates images for each corresponding RGBA image for future research. 

\vspace{-5pt}
\begin{table}[h]
    \begin{center}
    \begin{tabular}{ |l|ccc|}
    \hline
     \textbf{Split} & \textbf{Total} & \textbf{$D_{known}$} & \textbf{$D_{unknown}$} \\ 
     \hline \hline
     Vocab Size & 23 & 20 & 3 \\
    $D_{train}$ & 5094 & 3006 & 2088 \\
     $D_{test\_nc}$ & 1242 & 744 & 468 \\ 
   $D_{test\_v}$ & 989 & 580 & 409 \\ 
     \hline
    \end{tabular}
    \caption{Splits on RGBA Images}
    \label{data_split}
    \end{center}
    \vspace{-15pt}
\end{table}

To evaluate the novel composition capability of the methods, we reserved 9 learning attribute pairs exclusively in the Novel Composition Test set ($D_{test\_nc}$). The rest were assembled into the Train set ($D_{train}$) for word acquisition training. To evaluate models' abilities to continual learning, we split the vocabulary into two sets: a $\texttt{Known}$ vocabulary and an $\texttt{Unknown}$ vocabulary set, which leads to two datasets $D_{known}$ and $D_{unknown}$. The $D_{unknown}$ dataset includes images describable by at least one of the three attributes: [yellow, glass, torus\_knot], and the rest of the images are in $D_{known}$. Each training and testing dataset is broken down into $\texttt{Known}$ and $\texttt{Unknown}$ versions accordingly. More details about SOLA can be found in the Appendix.

Several existing datasets offer dense compositional attribute annotations that can be helpful for language grounding, such as MIT-States \cite{isola2015discovering}, UT-Zappos \cite{yu2014fine}, CUB \cite{wah2011caltech}, ShapeNet \cite{chang2015shapenet}, Visual Genome \cite{krishna2017visual}, and PACO \cite{ramanathan2023paco}. These datasets are great resources for scaling attribute concept learning, especially from noisy real world images, but are not designed to form clean structural alignment for comparative language acquisition. Our work took the baby step of progressive alignment \cite{hespos2020structure, kotovsky1996comparison, childerslanguage} by offering the model structured and denoised sets of inputs for easier structural comparison and efficient feature extraction. Following these works, we believe that equipping the model with a set of clean base knowledge can help it extend to messier inputs in the future. 

Other abstract datasets such as CLEVR \cite{johnson2017clevr} focus on diagnosing and probing the reasoning or interpretability of models through visual question and answering, and are not designed for language acquisition. Additionally, our dataset includes 2 more materials and 8 more shapes than CLEVR, providing a lot more variance and opportunities for vocabulary learning. We also diversify lighting, camera angles, and further object transformations in the variation test set for generalization and composition analysis. We introduce SOLA as it offers clean, grouped images for structured comparative learning. More detailed dataset comparisons can be found in Table \ref{dataset_compare}.

\vspace{-5pt}
\section{Method}
\vspace{-5pt}
\subsection{Comparative Learning}
\begin{figure*}
    \centering
    \includegraphics[width=\textwidth]{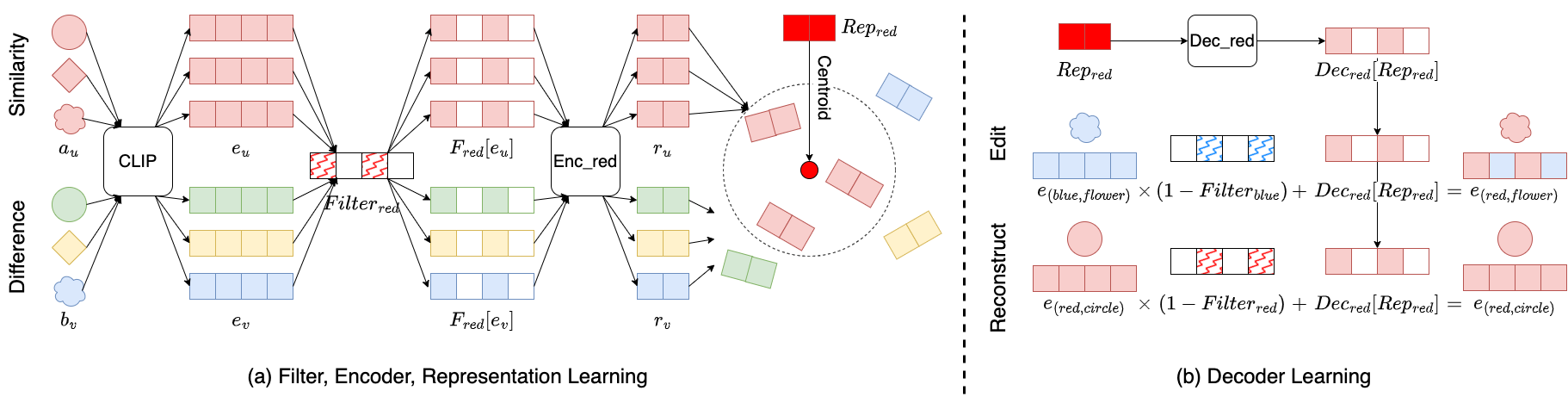}
    \caption{\textbf{Model Architecture and Learning Process:} (a) The encoder model learns to extract the shared features from the similarity batch, computes the centroid, and differentiates from the difference batch examples. (b) The decoder model learns to decompress the representation through image reconstruction and editing.}
    \label{ppline}
    \vspace{-15pt}
\end{figure*}

\textbf{Comparative Learning} is the process of finding the similarities and differences from a set of inputs. It is a general learning strategy that can be applied to different input modalities, sizes, and duration. The general formulation can be found below.

For each label/word/symbol $l_i$ in an unconstrained set $L = \{l_1, l_2, \cdots\}$, we first assemble a batch of samples $\mathcal{B}_s = \{a_1^{l_1}, a_2^{l_1}, \cdots, a_n^{l_1}\}$, that share the label $l_i$ for similarity learning, and a batch of samples $\mathcal{B}_d = \{b^{l_1}, \cdots, b^{l_j}, \cdots\}_{j\neq i}$ that cannot be described by $l_i$ for difference learning. The process of $\text{SIM}_{l_i}$ (Eq.\ref{eq:sim}) finds similarities across examples in $\mathcal{B}_s$, and extracts out its representation $\text{Rep}_{l_i}$. The process of $\text{DIFF}_{l_i}$ (Eq.\ref{eq:dif}) highlights the differences between $l_i$ and other non-compatible labels, and refines the representation $\text{Rep}_{l_i}$. Non-compatible labels are the ones that cannot be assigned to the same entity at the same time, e.g.(up, down). Comparable to the positive and negative batches in contrastive learning, these labels naturally occur through difference comparisons, and are organized by the supervisor. Both the computations and the representation are stored to map the label: \{$l_i$: [$\text{SIM}_{l_i}$, $\text{DIFF}_{l_i}$, $\text{Rep}_{l_i}$]\}. 

\vspace{-15pt}

\begin{flalign}\label{eq:sim}
    &\text{Rep}_{l_i} = \text{SIM}_{l_i}(\{a^{l_i} \in \mathcal{B}_s \})&\\
\label{eq:dif}
    &\text{Rep}_{l_i} = \text{DIFF}_{l_i}(a^{l_i}, \{b^{l} \in \mathcal{B}_d \})&
\end{flalign}
\vspace{-15pt}

In this work, we contextualize the method of comparative learning in word acquisition through a set of still visual inputs (Figure \ref{ppline}). For each concept, e.g. ``red'', we assemble a batch of images that share the word ``red'' for similarity training. We also assemble a batch of images that are of any other color (non-compatible) but ``red'' for difference refinement. We keep the rest of the attributes the same for better structural alignment.

As illustrated in Algorithm \ref{alg_enc} and Figure \ref{ppline}, given a batch of training samples (sim and diff) for word $l_i$: $\mathcal{B}=\{\mathcal{B}_s, \mathcal{B}_d\}$, we first take a shortcut by having each image $a_u$ go through a frozen pre-trained CLIP \cite{radford2021learning} image embedding as the starting point. This shortcut bypasses a few structural alignment steps, and encodes the raw images into the same 512 dimensions $e_u$ available for direct comparisons. The information denoising and attention establishment process is composed of two parts for each word $l_i$: the filter $\mathtt{F}_{l_i}$ and the encoder $\mathtt{Enc}_{l_i}$. The filter maintains a vector the same size as the embedding $e_u$, and computes the element-wise product of the two. It is a learning vector that masks the input embedding by selecting only the relevant dimensions that contributes to the word $l_i$. This masked embedding goes through two fully connected layers of $\mathtt{Enc}_{l_i}$ to output a condensed representation $r_{u}$. 

On top of learning the attention filtration process ($\mathtt{F}_{l_i}, \mathtt{Enc}_{l_i}$), we then calculate the centroid of all the sample representations $r_{u}$ from the similarity batch $\mathcal{B}_s$ as the condensed representation $\text{Rep}_{l_i}$ for $l_i$. For difference learning, we have all the $\mathcal{B}_d$ samples to go through the same filtration and encoding process for word $l_i$. Since none of them can be described by the word $l_i$, the output should be nothing like $\text{Rep}_{l_i}$. Therefore, the loss function pushes the distance between each sim batch sample and the centroid close, and pushes the diff batch sample representations apart from the centroid. 

This process filters out and abstracts the shared representations of $l_i$, and differentiates it from other non-compatible words. It jointly trains the filter $\mathtt{F}_{l_i}$, the encoder $\mathtt{Enc}_{l_i}$, and the representation $\text{Rep}_{l_i}$. We store the mapping \{$l_i$: [$\mathtt{F}_{l_i}$, $\mathtt{Enc}_{l_i}$, $\text{Rep}_{l_i}$]\} for each word in memory for later use. 


\vspace{-5pt}
\begin{algorithm}[!ht]
\DontPrintSemicolon
\For{Sim and Diff data batches: $\{\mathcal{B}_s, \mathcal{B}_d\} $}
{
      \textcolor{lightgray}{\tcp{Similarity Learning}}
      \For{$a_u \in \mathcal{B}_s$}
        {
            $e_{u} = \mathtt{CLIP\_emb}[a_u]$\\
            $r_{u} = \mathtt{Enc}_{l_i}[\mathtt{F}_{l_i}(e_{u})]$
        }
    $\text{Rep}_{l_i} = \mathtt{Centroid}[\{r_{u}\}_{u\in \{1, \cdots, n\}}]$\\
    \textcolor{lightgray}{\tcp{Difference Learning}}
        \For{$b_v \in \mathcal{B}_d$}
        {
            $e_{v} = \mathtt{CLIP\_emb}[b_v]$\\
            $r_{v} = \mathtt{Enc}_{l_i}[\mathtt{F}_{l_i}(e_{v})]$
        }
    \textcolor{lightgray}{\tcp{Loss}}
    $\text{loss}_s = \sum_{u} \mathtt{Dist} [r_{u}, \text{Rep}_{l_i}]$\\
    $\text{loss}_d = \sum_{v} \mathtt{Dist} [r_{v}, \text{Rep}_{l_i}]$\\
    $\text{loss} = (\text{loss}_s)^2 + (1-\text{loss}_d)^2$\\
    Backpropagate and Optimize
}
\KwOutput{\{$l_i$: [$\mathtt{F}_{l_i}$, $\mathtt{Enc}_{l_i}$, $\text{Rep}_{l_i}$]\}}
\caption{Comparative Learning-Word $l_i$}
\label{alg_enc}
\end{algorithm}
\vspace{-5pt}

\subsection{Generative Decoder Learning}
Due to input filtration, the dimensions of the condensed word representations come from selective, word-specific subsets of the original 512 dimensions of $e$. They are, therefore, not aligned in the same space across different words and cannot be used for direct interactions. To allow compositional reasoning with all the words and their grounded representations, we trained a decoder (Figure \ref{ppline}) to revert the condensed representations back to the same space as the CLIP embedding $e$. 

To train the decoder $\mathtt{Dec}_{l_p}$ for word $p$, we adopted two strategies in parallel: \underline{Editing} and \underline{Reconstruction} (Figure \ref{ppline}). About editing, given an image of a (\textcolor{blue}{blue}, flower), for example, if we filter out \textcolor{blue}{blue} add \textcolor{red}{red}, we should get an image of a (\textcolor{red}{red}, flower). Following this logic as in Eq. \ref{eq_edit}, we mask out feature $q$ from input embedding $e_q$ by multiplying the opposite of filter $q$: $(1 - \mathtt{F}_{l_q})$. We then add back the decoded ($\mathtt{Dec}_{l_p}$) representation of $\text{Rep}_{l_p}$ for word $p$. Both the filter $\mathtt{F}_{l_q}$ and the representation $\text{Rep}_{l_p}$ were trained in the previous step and frozen. The output ($\text{out}\_{q2p}$) aims to resemble the embedding of $e_p$. Similarly, for reconstruction as in Eq. \ref{eq_recon}, if we filter out feature $p$ from input embedding $e_p$ and add back the decoded representation of $\text{Rep}_{l_p}$, we should get the original embedding of $e_p$. Both passes are trained jointly to learn the decoder of $p$ (Eq. \ref{eq_declos}).

Each decoder is stored together in the mapping \{$l$: [$\mathtt{F}_{l}$, $\mathtt{Enc}_{l}$, $\mathtt{Dec}_{l}$, $\text{Rep}_{l}$]\}. The decoded representations open the door for zero-shot compositional comprehension, generation, and reasoning. For illustration purpose, we also trained a small image generator that upsamples the CLIP embedding back to an RGB image. Details about the models and training can be found in the Appendix.


\vspace{-20pt}
\begin{flalign}
     \label{eq_edit}
    &\text{out}\_{q2p} = e_{q} (1 - \mathtt{F}_{l_q}) + \mathtt{Dec}_{l_p}[\text{Rep}_{l_p}]&\\
    \label{eq_recon}
    &\text{out}\_{p2p} = e_{p} (1 - \mathtt{F}_{l_p}) + \mathtt{Dec}_{l_p}[\text{Rep}_{l_p}] &\\
     \label{eq_declos}
    &\text{loss} = \mathtt{Dist} [e_{p}, \text{out}\_{q2p}] + \mathtt{Dist} [e_{p}, \text{out}\_{p2p}] &
    \vspace{-20pt}
\end{flalign}
\vspace{-30pt}

\section{Experiments}
\vspace{-5pt}
With the training described above, each word will have a mapping
 \{$l$: [$\mathtt{F}_{l}$, $\mathtt{Enc}_{l}$, $\mathtt{Dec}_{l}$, $\text{Rep}_{l}$]\} stored in the memory. 
 These acquired word representations can be used during inference time for downstream tasks. In this section, we evaluate these representations on several tasks that test models' robustness, generalizability, flexibility, and ability to continual learning.

\subsection{Multi-Attribute Recognition} \label{sec_marecg}

\begin{figure}[]
    \includegraphics[width=0.49\textwidth]{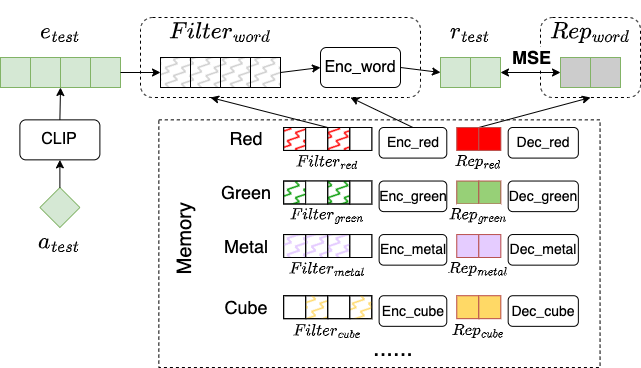}
  \caption{Multi-Attribute Recognition Inference}
  \label{infer}
  \vspace{-10pt}
\end{figure}

In this task, the models are challenged with zero-shot recognition of all the attributes (color, shape, material) of a given test image $a$ under two evaluation settings: (1) {\em Novel composition setting} where the image with a combination of attributes which is not seen during training (i.e., $a \in D_{test\_nc}$); and (2) {\em Noisy setting} where the images were injected with noise in the variation test set ($a \in D_{test\_v}$). The models were trained on the training data ($D_{train}$). For each test image (Figure \ref{infer}), we go through the memory, apply the corresponding filter and encoder of each word to the input embedding, and picked the top 3 words with the shortest mean squared error (MSE) between the learned word representation and image embedding. 



 We took the essence of several zero-shot compositional learning methods such as \cite{zscl_li, zscl_anwaar, zscl_Mancini}, and implemented them as variations of the CLIP model for a better experiment control and fairer comparison. 
 More specifically, we compare our method with the following baselines: \\
\textbf{CLIP Zero Shot} computes the highest matching words for each test image. We experimented with different prompts, and reported the highest performances using the prompt ``a photo of a {x}''.\\
\textbf{CLIP Contrastive} adds two fully connected layers to the image encoder, and fine tune the model on the same training dataset with a contrastive loss.\\
\textbf{CLIP Linear} also adds two fully connected layers to the image encoder, but with an output dimension of the vocabulary size. It predicts 1s for all corresponding word dimensions, and 0s otherwise. This method is subject to a fixed vocabulary size, and can be hard to expand to new concepts.\\
\textbf{CLIP Multi-Attr} finetunes two fully connected layers out of the image encoder for \underline{each} word, and predicts 1s and 0s based on its confidence measured by the word-image matchability (i.e., similarity). 

\begin{figure}[]
  \begin{subfigure}[b]{0.49\textwidth}
    \includegraphics[width=\textwidth, center]{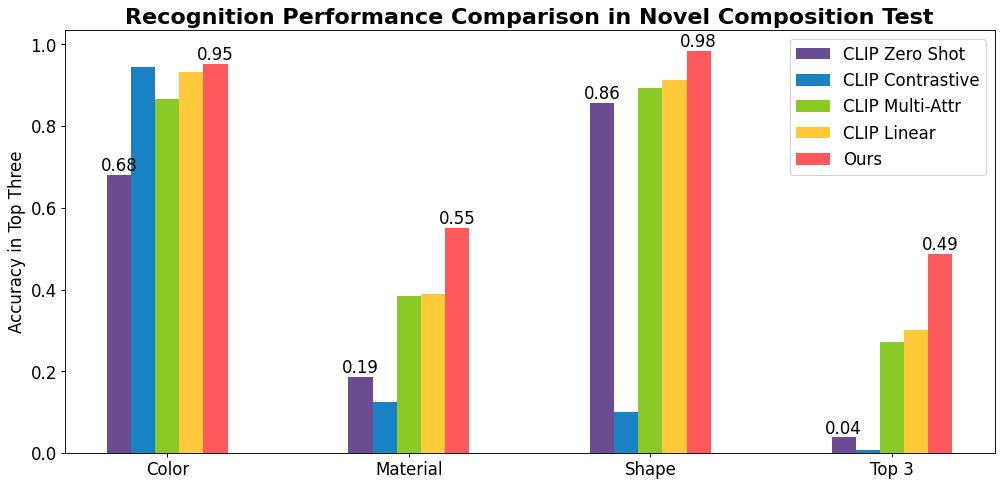}
    \caption{Novel Composition Test}
    \label{recog_nt}
  \end{subfigure}
  \hfill
  \begin{subfigure}[b]{0.49\textwidth}
    \includegraphics[width=\textwidth, center]{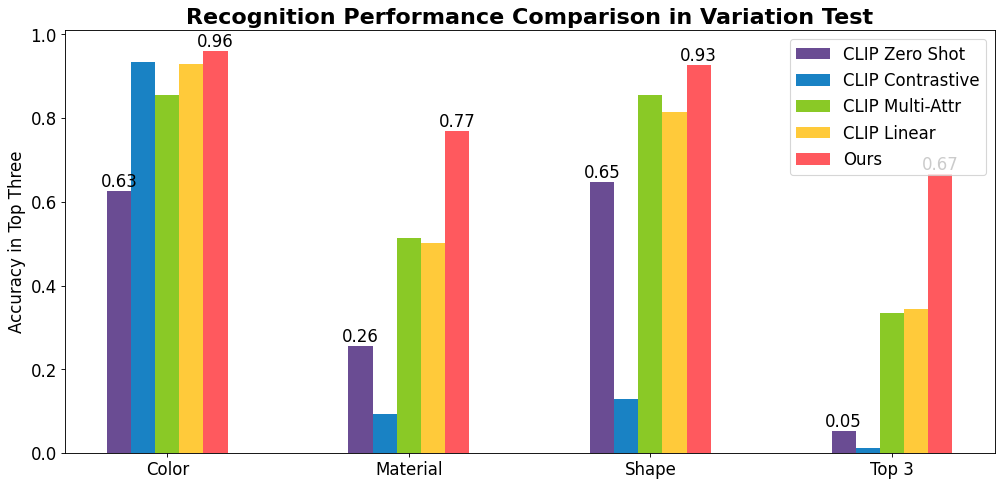}
    \caption{Variation Test}
    \label{recog_t}
  \end{subfigure}
  \caption{\textbf{Multi-Attribute Recognition Performance Comparison}: The percentage of each ground truth attribute (color, material, shape, or all 3) being among the top 3 model predictions.}
  \label{recog}
  \vspace{-20pt}
\end{figure}

The performance of all the methods over two test datasets can be found in Figure \ref{recog}. For each image, we evaluate whether its corresponding color, material, shape, or all three of them are among the top 3 selected words. 


It is observed that our method consistantly outperforms all baselines across two test datasets and four categories. CLIP Zero Shot showed decent baseline performance on the multi-attribute recognition task, as this model was pre-trained on massive weakly linked language-image pairs. However, our model and the finetuned models are able to surpass this baseline with a significant margin. CLIP Contrastive overfits to the color features, mainly guessing colors in its top three resulting in high color performance but lagging behind in all other attributes. CLIP Linear and CLIP Multi-Attr showed an improved performance compared to CLIP Zero Shot, but couldn't catch up with our method. 

Among the 3 attributes, the material attribute was the hardest to learn for all the methods. Humans generally learn textures through touching, tapping an object for sound, and other sensors so a visual input alone may not be sufficient to grasp the meaning of materials, especially under a dim light. However, our method was still able to lead in performance on materials, which consequently also increased the accuracy for all top 3. This is likely because our model is able to pay attention to specific aspects (e.g. light reflection, transparency) of the images better through explicit comparisons.


\vspace{-5pt}
\subsection{Continual Word Acquisition}

\begin{figure}[]
    \includegraphics[width=0.49\textwidth]{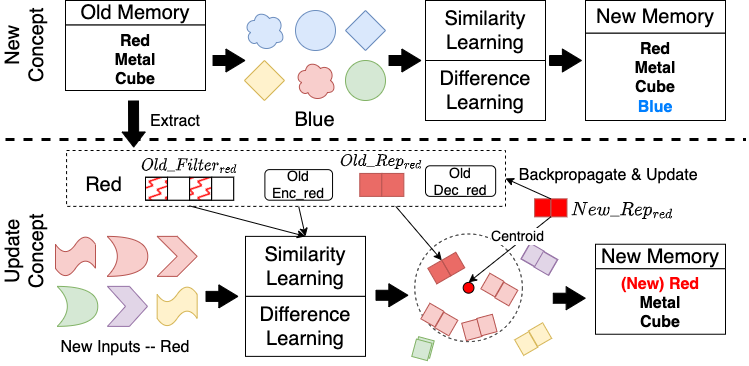}
    \caption{\textbf{Continual Learning}: 1) New concepts can be continually added to memory using the same method in Figure \ref{ppline}; 2) Existing concepts can be updated and refined as more samples are introduced.}
    \label{cont_ppl}
    \vspace{-15pt}
\end{figure}

We investigated models' capability to continually acquire new words on the same multi-attribute recognition task in comparison with the models mentioned in Section \ref{sec_marecg}. 
As mentioned in Section \ref{sec_data}, we split all the training and testing datasets into two parts based on the vocabulary ($D_{known}$, $D_{unknown}$). The $D_{known}$ datasets include 20 words, and the $D_{unknown}$ datasets include an additional 3 new words for continual word acquisition and evaluation. Any image that shares at least one of the 3 new words is part of $D_{unknown}$.

Our model conducts continual learning in two ways (Figure \ref{cont_ppl}) it can learn new concepts using the exact same way as described in Figure \ref{ppline}, and add the word-representation mapping to the memory; 2) It can also update and refine the learned concepts, whenever new samples are available. More specifically, we extract the relevant \{$l$: [$\mathtt{F}_{l}$, $\mathtt{Enc}_{l}$, $\mathtt{Dec}_{l}$, $\text{Rep}_{l}$]\} from the memory for word $l$. The new samples go through similarity and difference learning with the old $\mathtt{F}_{l}$ and $\mathtt{Enc}_{l}$ to get a batch of condensed representation $\{r\}$'s. Together with the old $\text{Rep}_{l}$, we can calculate a new centroid with these embeddings, and a loss. Through backpropogation and training, the new centroid will be the refined $\text{Rep}_{l}$, and both the encoder and filter are updated in the memory for word $l$.

\begin{figure}[t]
  \begin{subfigure}[]{0.49\textwidth}
    \includegraphics[width=\textwidth, center]{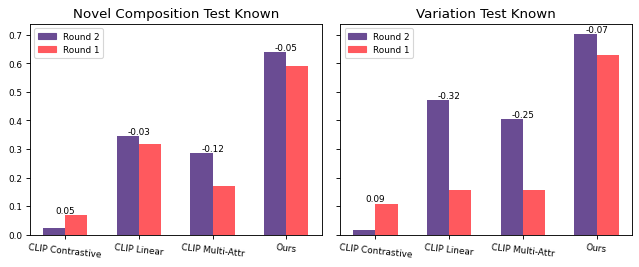}
    \caption{Catastrophic Forgetting on Test Datasets $D_{known}$}
    \label{cont_1}
  \end{subfigure}
  \hfill
  \begin{subfigure}[]{0.49\textwidth}
    \includegraphics[width=\textwidth, center]{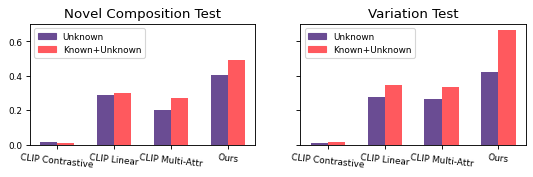}
    \caption{Data Efficiency in Round 2: Trained on $D_{unknown}$ versus $D_{known}$+$D_{unknown}$}
    \label{cont_2}
  \end{subfigure}
  \vspace{-5pt}
  \caption{\textbf{Continual Word Acquisition Evaluation}: Accuracy of model's top 3 predictions being all ground truth attributes.}
\label{cont_eval}
\vspace{-20pt}
\end{figure}

We first evaluate the severity of catastrophic forgetting of the methods (Figure \ref{cont_1}). In Round 1, the models were trained on $D_{known}$ datasets, and the $D_{unknown}$ sets in Round 2. We evaluate the accuracy of the models on two $D_{known}$ test sets by computing the percentage of models' top 3 predictions all being the ground truth attributes.

For CLIP Contrastive, we increased the vocab size for Round 2 training. For CLIP Multi-Attr and our method, we introduced additional models for each new concept. The CLIP Linear model was the hardest to grow as the output dimension was fixed to the previous vocab size. We initialized the first linear weights with the learned weights in Round 1, and had to retrain the model in Round 2.

In Figure \ref{cont_1}, except for the CLIP Contrastive model, most models suffered from catastrophic forgetting between Round 1 and Round 2. Our method had a mild performance decrease as more vocab was introduced. This is likely due to the top 3 label selection competitions among increasing vocab size. CLIP Linear and CLIP Multi-Attr suffered severe catastrophic forgetting on the Variation Test $D_{known}$ set, likely due to lack of generalizability. 

We also evaluated the continual training data efficiency for different models. During Round 2, we compare how much replay data would the models need in order to achieve a decent performance by training them on either the $D_{unknown}$ datasets only (new lessons) or both the $D_{known}$+$D_{unknown}$ datasets (new and old lessons). Round 2 trained on only $D_{unknown}$ receives significantly less data, and does not require reviewing old lessons. 

In Figure \ref{cont_2}, when trained only with the $D_{unknown}$ set, our method had already outperformed all other methods even compared to their performances when trained with both $D_{known}$+$D_{unknown}$ datasets. When more data was available, our method was able to improve performance even further on identifying all attributes.

These results showed early signs of efficient continual learning capabilities and resistance against catastrophic forgetting. Unlike discriminative methods such as CLIP Linear, which has a fixed output dimension based on the vocab size, our method is a lot more flexible to grow for new concepts, and achieved higher performance without the need to review old concepts. Further investigations are needed for larger scale evaluation.



\vspace{-5pt}
\subsection{Compositional Imagination and Reasoning}

Another way of evaluating acquired words is through compositional imagination and reasoning given words. With the stored meaning representations, we will be able to flexibly compose different meanings together for reasoning, imagination, simulation, and language understanding. We evaluate this capability in two use cases: composition reasoning and generation. 

Most traditional multimodal methods, such as the ones in Section \ref{sec_marecg} only focus on learning a feature extractor given an image. They do not store a grounded representation for each word for reasoning or generation. We therefore, have to turn to the text embedding part of CLIP for comparison as they were trained to be in the same space as the image embeddings. Text embeddings have been shown to carry grounded semantic meanings through high resolution image generations, but also have been found to struggle at grounding certain attributes \cite{imagen, dalle2}.

In this section, we compare our method to \textbf{CLIP Zero Shot} and \textbf{CLIP Finetune} on the following tasks. We use the text embedding of both methods to do image editing and compositional generation. For CLIP Finetune, we added two fully connected layers on the text embedding and fintuned with our $D\_train$ dataset.



\subsubsection*{Composition Generation}
Without any given images, humans are able to build mental representations of objects given linguistic descriptions. These representations are built upon abstract word associated features, and can be flexibly composed and manipulated as more features are added. Unlike previous works that emphasize on high resolution image generation, we focus on building compositional mental representations that can be used for downstream reasoning.

\begin{figure}[]
    \includegraphics[width=0.49\textwidth]{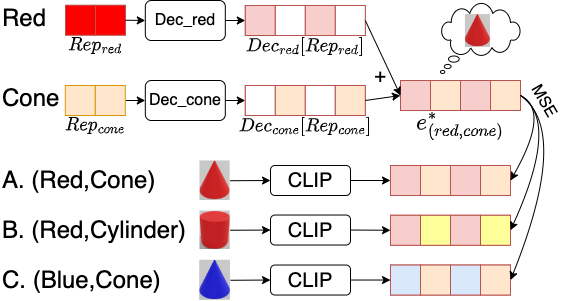}
    \caption{Novel Composition Imagination Multiple Choice Evaluation}
    \label{mc_ppl}
    \vspace{-15pt}
\end{figure}

Quantitatively, we evaluate the fidelity of a pair of concept composition through a multiple choice task. Given any two random compatible words, e.g. (\textcolor{red}{red}, \textcolor{blue}{cone}), and the CLIP embedding of two hard distractors (each sharing at least one attribute as the original pair, e.g. A. (\textcolor{red}{Red}, \textcolor{blue}{Cone}), B. (\textcolor{red}{Red}, Cylinder), C. (Blue, \textcolor{blue}{Cone})), we challenge the models to generate a mental representation such that it is closest to the correct image embedding. Each choice is a randomly selected image with the two specified features. As shown in Figure \ref{mc_ppl}, for example, we decode representations of both ``red'' and ``cone'' and then add the two resulting vectors to create our ``mental image'' embedding of a red cone. The multiple choice embedding with the smallest MSE is chosen to be the desired attribute composition imagination. 

\begin{table}[h]
\small
\centering
\begin{tabular} 
{>{\centering\arraybackslash}m{0.9cm}|>{\centering\arraybackslash}m{1.1cm}|>{\centering\arraybackslash}m{1.1cm}|>{\centering\arraybackslash}m{1.1cm}|>{\centering\arraybackslash}m{1.1cm}}
& \textbf{C+M} & \textbf{C+S} &\textbf{M+S} & \textbf{All}\\	
\hline  \hline
Zero Shot & 53.8$\pm$3.2 & 74.5$\pm$4.7 & 51.0$\pm$6.1 & 59.8$\pm$3.3\\
Finetune & 53.9$\pm$4.4 & 75.3$\pm$4.2 & 57.9$\pm$6.2 & 62.4$\pm$3.1\\
Ours & \textbf{58.1}$\pm$\textbf{5.1} & \textbf{82.3}$\pm$\textbf{4.4}& \textbf{59.7}$\pm$\textbf{3.3}& \textbf{66.7}$\pm$\textbf{2.6}\\ 
\end{tabular}
\caption{Composition Generation Multiple Choice Task Performance. \textbf{C}=Color, \textbf{M}=Material, \textbf{S}=Shape.}
\label{gen_comb}
\vspace{-10pt}
\end{table}


The performance can be found in Table \ref{gen_comb}. Over 15 runs of 100 randomly selected questions each, our model is able to outperform the text embedding composition of both CLIP Zero Shot and CLIP Text Finetune. Among those, the Color+Shape combo is the easiest one to assemble, likely due to the challenges of learning the material features for the other two combinations. Our method is better at extracting different aspects of the inputs, and learn to pin down the exact word meanings through efficient similarity and difference comparisons.

\begin{figure}[!tbp]
    \includegraphics[width=0.49\textwidth]{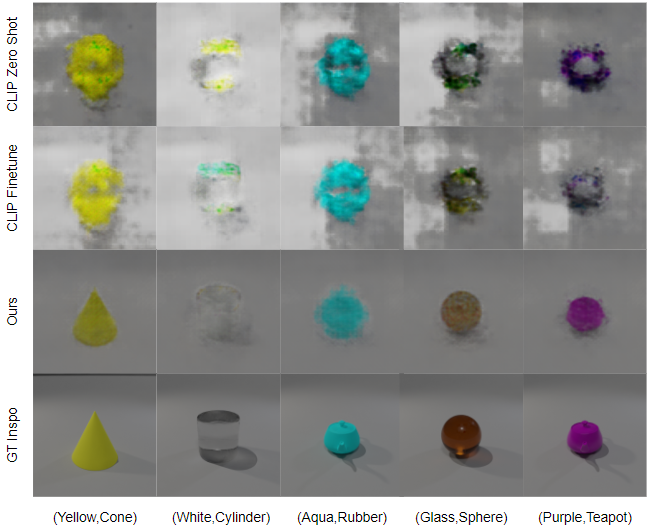}
    \caption{\textbf{Novel Composition Generation}: Our method is able to accurately reflect the word, and zero-shot compose mental images with novel feature combinations.}
    \label{comp}
    \vspace{-20pt}
\end{figure}

Qualitatively, we also generated the representations of the novel combinations in our testing data (Figure \ref{comp}), and see how visually close they are to the GT pictures.  The visualization shows that CLIP Zero Shot and CLIP Finetune both struggle at representing the precise definition of some shape concepts, such as ``Cone'', ``Sphere'', and ``Teapot'', but are good at embedding color concepts. The last row serves as an inspiration of possible ground truth images given a pair of concepts. 

 \vspace{-5pt}

\vspace{-1pt}
\subsubsection*{Composition Reasoning}
Another simple compositional reasoning task on different object features is to do `arithmetic' with them. For example, a (\textcolor{red}{red}, cone) - \textcolor{red}{red} + \textcolor{blue}{blue}= (\textcolor{blue}{blue}, cone). With the Decoder training in Figure \ref{ppline}, we can flexibly edit a given image to a desired feature. In this section, given an input image, and a random pair of attribute switching, we qualitatively evaluate if the edited image resembles the desired attribute while keeping all other features the same.

\begin{figure}[]
    \includegraphics[width=0.49\textwidth, center, scale=0.75]{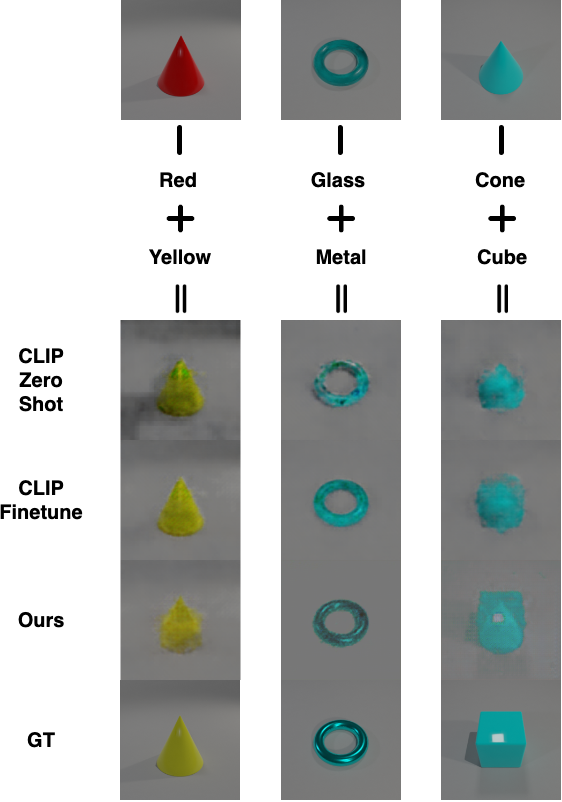}
  \caption{\textbf{Composition Reasoning}: Our method can perform image feature editing given grounded word representations, and performs better than CLIP Zero Shot and CLIP Finetune on material and shape editing.}
  \label{reason}
  \vspace{-15pt}
\end{figure}

Figure \ref{reason} shows three qualitative examples on feature switching over color, material, and shape, compared to CLIP Zero Shot and CLIP Finetune. It is observed that CLIP trained text embeddings excel at extracting color related concepts, but struggle at material and shape. This could be due to its unfamiliarity with the specific material and shape words that we use in our dataset, whereas color concepts are more universal and are easier to learn from pixels. 
Finetuning helps improve the performance, but still lagged behind our method. More qualitative examples can be found in the Appendix.

\vspace{-5pt}
\section{Conclusion}
\vspace{-5pt}
In this work, we took a human inspired approach to acquire multi-attribute concepts. We define the acquisition of word as learning both an information filtration process, and a representation-symbol mapping. We mimic the classroom setting, constructed a small clean dataset SOLA for efficient comparative and continual learning. We evaluated the learned representations in multi-attribute recognition, compositional simulation and reasoning tasks.  
Our experiment results outperformed CLIP variations in controlled settings, and showed early signs of a promising new method for continual grounded word acquisition through comparative learning.

\clearpage
\section*{Limitations}
As exciting as this work is, it does have several limitations and a lot of opportunities for future improvement. 

\textbf{How to scale?} We demonstrated our method in a highly constrained environment with very limited concepts, whereas humans are able to pick up new concepts in the noisy world with few shots. How could these representations learned in a clean environment be useful in real world? Would comparative learning still be useful outside of the classroom? We followed the baby step of progressive alignment and hoping that establishing a set of clean base knowledge, can ease the acquisition of future more complex concepts through comparisons with existing knowledge, analogy and hierarchical abstraction. This hypothesis remains to be investigated in the future.

\textbf{What about other words?} Some concepts can be learned through just visual inputs, like color, whereas other concepts require grounding through different sensory types or modalities, like ``hot'', ``loud'' and ``fast''. Even more concepts are built upon existing words through abstraction and generalization, e.g. ``philosophy'', ``momentum''. Comparisons can still be used to ground these words, but input to these comparisons could vary from data modalities to computation methods, to abstract representations. We leave these for future work.

\textbf{How to put words into sentences?} This work only focused on the grounding of individual words into visual representations, whereas sentence syntax, grammar, and article structure are yet to be learned. For future work, we could treat language as its own modality, and learn the structure through comparisons as well. Just like in an elementary linguistic class, a teacher would list out several examples ``I shower''/``You shower''/``He shower''. Humans can learn grammar through what's changing and what's constant. This could be an interesting next step to look into.

\textbf{Who can offer the supervision?} As mentioned at the beginning, human language acquisition is a highly supervised learning process. Babies are rarely inventing new words but learning how adults label objects through generations of conventions. A classroom setting with highly structured curriculum and clean dataset takes a lot of curriculum design and heavy annotation. This is the cost that humans are willing to spend in order to educate human children from kindergarten to college. Maybe it is a fair price that we have to pay in order for artificial intelligence to learn what we want them to learn. 

About the current work itself, there are several constraints that we are limited to.

First of all, due to limited computation resources and data size, we had to take a shortcut by using a pre-trained CLIP embedding as a starting point for our models. In theory, we could and would love to train our models from scratch, just like how a new born would learn their first language. A dataset like Toys-200  \cite{toy200} could mimic the process of babies interacting with the objects, get a 360 view and help build 3D mental representations. 

Second of all, like many other continual learning methods, an unbounded memory space is an unrealistic assumption. As more concepts are learned, the memory space would grow fast, so as the search time. An interesting next step could be to re-organize the memory according to the association distances and hierarchical structures.   

Lastly, our work aims at proposing a novel language acquisition definition and the comparative continual learning method. We used somewhat simple model architecture and image generation models for proof-of-concept demonstration on the method. More sophisticated model architecture and training can be switched for different input modalities and applications.

Listed above are several major limitations and future directions based on current work. We are more than happy to take constructive suggestions and criticism to help improve this and future works.

\section*{Ethics Statement}
This work took the human inspired approach to learn word acquisition for artificial intelligent agents. We generated a small clean dataset using the open-source simulation software Kubric, which was designed for semi-realistic image/video synthesis. All of our training was done on a single machine with 8GB GPU and an Interl i9 processor, with very limited environmental cost. This work does not involve human subject, nor can be can be used to directly interact with humans. 


\section*{Acknowledgements}
This work was supported in part by NSF IIS-1949634 and DARPA PTG program HR00112220003. The authors would like to thank
the anonymous reviewers for their valuable comments and suggestions.

\clearpage
\bibliography{anthology,custom}
\bibliographystyle{acl_natbib}

\clearpage
\appendix

\section{Dataset SOLA} \label{appx_data}

\begin{figure}[]
  \begin{subfigure}[b]{0.49\textwidth}
    \includegraphics[width=\textwidth, center]{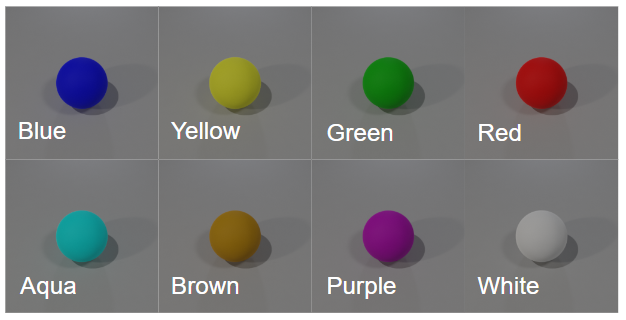}
    \caption{8 Colors}
    \label{dt_C}
  \end{subfigure}
  \hfill
  \begin{subfigure}[b]{0.49\textwidth}
    \includegraphics[width=\textwidth, center]{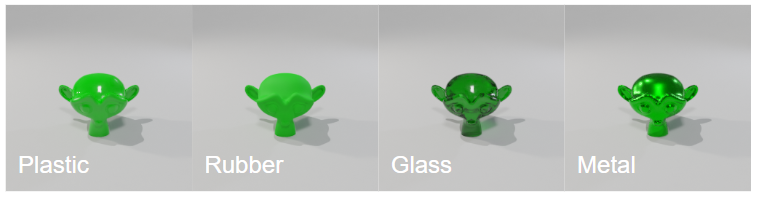}
    \caption{4 Materials}
    \label{dt_m}
  \end{subfigure}
  \hfill
  \begin{subfigure}[b]{0.49\textwidth}
    \includegraphics[width=\textwidth, center]{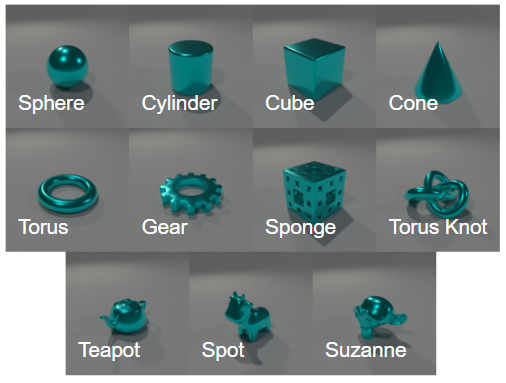}
    \caption{11 Shapes}
    \label{dt_s}
  \end{subfigure}
  \caption{SOLA Learning Attributes}
  \label{dt_la}
    \vspace{-15pt}
\end{figure}

\begin{figure}[]
  \begin{subfigure}[b]{0.49\textwidth}
    \includegraphics[width=\textwidth, center]{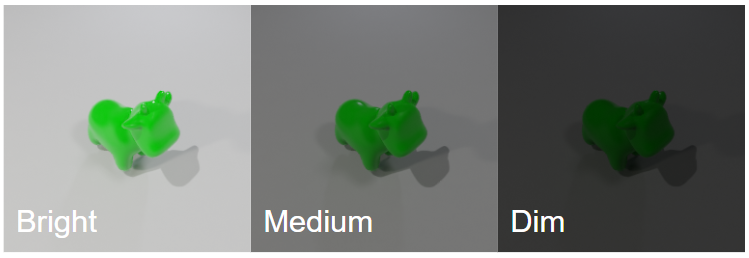}
    \caption{3 Lighting Conditions}
    \label{dt_b}
  \end{subfigure}
  \hfill
  \begin{subfigure}[b]{0.49\textwidth}
    \includegraphics[width=\textwidth, center]{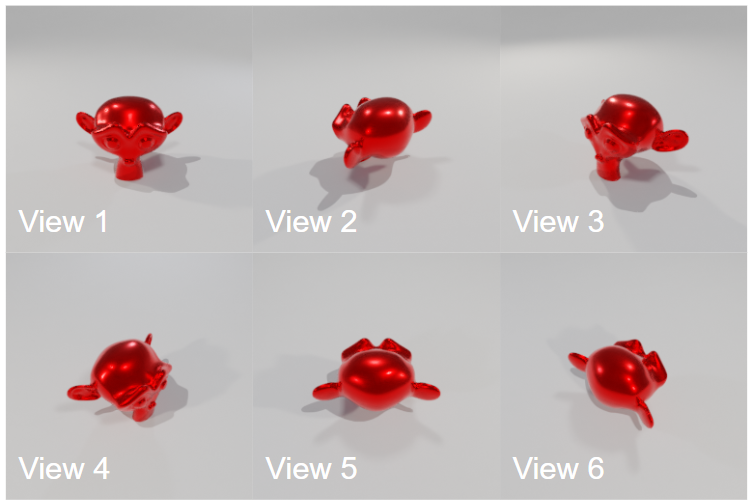}
    \caption{6 Views}
    \label{dt_v}
  \end{subfigure}
  \caption{SOLA Environment Augmentation}
  \label{dt_ca}
  \vspace{-15pt}
\end{figure}

\begin{figure}[]
  \begin{subfigure}[b]{0.49\textwidth}
    \includegraphics[width=\textwidth, center]{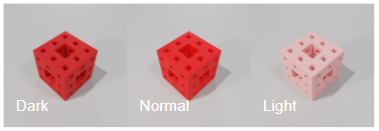}
    \caption{3 Shades per Color}
    \label{dt_shade}
  \end{subfigure}
  \hfill
  \begin{subfigure}[b]{0.49\textwidth}
    \includegraphics[width=\textwidth, center]{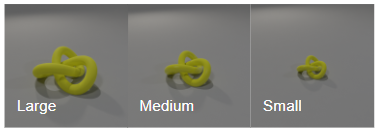}
    \caption{3 Sizes}
    \label{dt_size}
  \end{subfigure}
  \hfill
  \begin{subfigure}[b]{0.49\textwidth}
    \includegraphics[width=\textwidth, center]{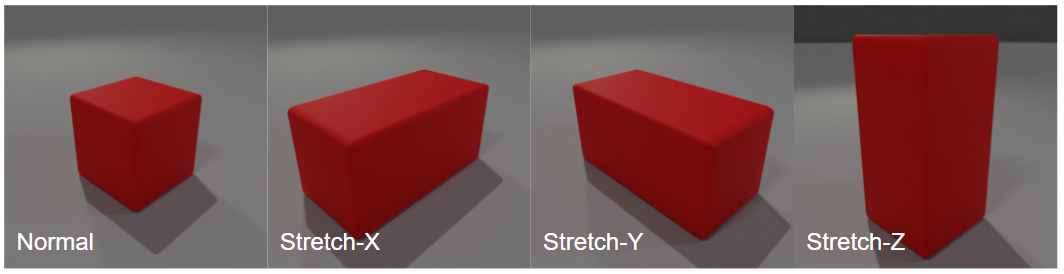}
    \caption{4 Stretches}
    \label{dt_stretch}
  \end{subfigure}
  \caption{SOLA Variations}
  \label{dt_va}
  \vspace{-10pt}
\end{figure}

Here is a detailed description of the Simulated Objects for Language Acquisition (SOLA) dataset:\\
\underline{Learning Attributes} (Figure \ref{dt_la}):
\begin{enumerate}
    \item Color: 8
    \item Material: 4
    \item Shape: 11
\end{enumerate}
\underline{Changing Attributes} (Figure \ref{dt_ca}):
\begin{enumerate}
    \item Lighting: 3
    \item Camera Angle: 6
\end{enumerate}
\underline{Variation Attributes} (Figure \ref{dt_va}):
\begin{enumerate}
    \item Shade: 3
    \item Size: 3
    \item Stretch: 4
\end{enumerate}
\underline{Image Types}:
\begin{enumerate}
    \item RGBA
    \item Depth
    \item Surface Normal
    \item Segmentation
    \item Object Coordinates
Coordinates
\end{enumerate}

This amounts to 7325 RGBA images in total, with 6336 originals and 989 with variations. A training and testing split can be found in Table \ref{data_split}. The original image set was first broke down into Novel Composition Training and Novel Composition Testing. 9 pairs of attributes are:
\begin{enumerate}
    \item (yellow, cone)
    \item (green, metal)
    \item (plastic, cube)
    \item (purple, teapot)
    \item (red metal)
    \item (glass, torus\_knot)
    \item (white, cylinder)
    \item (aqua, rubber)
    \item (glass, sphere)
\end{enumerate}

For continual learning evaluation, we split the vocabulary into the following two sets. Any images associated at least one of the concepts in $D_{unknown}$ are assembled into $D_{unknown}$ train/test datasets, and the rest in $D_{known}$. The number of samples in each split can be found in Table \ref{data_split}.\\
$\texttt{Known}$ = [brown, green, blue, aqua, purple, red, white,
      rubber, material, plastic, 
      cube, cylinder, sphere, cone, torus, gear, sponge, spot, teapot, suzzane]\\
$\texttt{Unknown}$ = [yellow, glass, torus\_knot]


\section{Model Architecture and Training} \label{appx_model}
For the encoder training, we used the pretrained CLIP image encoder (frozen) to embed the input images, going through a filter of 512 dimensions, and two fully connected layers with a hidden dimension of 128 and latent dimension of 16. Each round is trained on a similarity batch and a difference batch of size 128 each. The training moves on to the next concept when the loss went down below 0.008 or hit 200 rounds. The whole vocabulary was trained for 5 epochs with a learning rate of $1e^{-3}$.

For the decoder training, we froze the weights of the filter and the pre-trained representations from the previous step, and trained four fully connected layers with a dimension upsampling $16 \rightarrow 64 \rightarrow 64 \rightarrow 96 \rightarrow 512$ with a dropout rate of 0.2. Each concept was trained for 100 round with a batch size of 128. The whole vocabulary was trained for 5 epochs with a learning rate of $1e^{-3}$.

For comparisons, CLIP Contrastive embedded both image inputs, and text inputs. The image embeddings went through two fully connected layers with a hidden dimension of 128 and output dimension of the vocabulary size. CLIP Linear trained two fully connected layers on top of the image embeddings with a hidden dimension of 128 and output dimension of the vocabulary size. CLIP Multi-Attr did the same thing for each word, and the output dimension was 1 over softmax predictions. CLIP Text Finetune trained two fully connected layers on top of the text embeddings, with an input \& output dimensions of 512, and hidden dimension of 66. We tried to keep all the model architecture relative the same or having similar number of parameters for a fair comparison. The models were each trained for 50 epochs with a learning rate of $1e^{-3}$.

The small image generator contains 5 up-sampling convolution layers with dimensions going from (512,1) to (3,224,224). The number of channels are [128, 64, 32, 16, 3]. We trained 100 epochs on our original dataset with a learning rate or $1e^{-3}$.

All experiments done on a single NVIDIA(R) GeForce(R) RTX 2070 SUPER(TM) 8GB GDDR6 and 10th Gen Intel(R) Core(TM) i9-10900K processor.

\section{SOLA and Other Dataset Comparisons}
\begin{table*}
\small
\begin{center}
\begin{tabular}{|>{\centering\arraybackslash}p{2cm}|>{\centering\arraybackslash}m{1cm}|>{\centering\arraybackslash}m{2cm}|>{\centering\arraybackslash}m{3cm}|>{\centering\arraybackslash}m{3cm}|>{\centering\arraybackslash}m{2cm}|}
\hline
 \textbf{Dataset} & \textbf{Size} & \textbf{Language} & \textbf{Image Type(s)} & \textbf{Purpose} & \textbf{Structural Alignment} \\ 
 \hline  \hline
 CUB & 11.8k & Sentences & rgb, bbox & fine-grain classification & No \\
 UT-Zappos & 50k & Words & rgb & attribute comparison & No \\
 ShapeNet & 51k & Words & 3D & 3D shapes & No \\
 MIT-States & 53k & Words & rgb & state transformation & No \\
 PACO & 81.5k & Words & rgb, seg & part segmentation & No \\
 CLEVR & 100k & Sentences & rgb & reasoning diagnosis & No \\
 Visual Genome & 108k & Sentences & rgb, bbox & question answering & No \\ \hline
 SOLA & 36.6k & Words & rgba, depth, seg, surf form, obj cords & comparative acquisition & Yes \\ \hline
\end{tabular}
\caption{Multimodal Dataset Comparisons}
\label{dataset_compare}
\end{center}
\vspace{-20pt}
\end{table*}

\clearpage
\begin{figure*}[]
  \begin{subfigure}[b]{0.9\textwidth}
    \includegraphics[width=\textwidth, center]{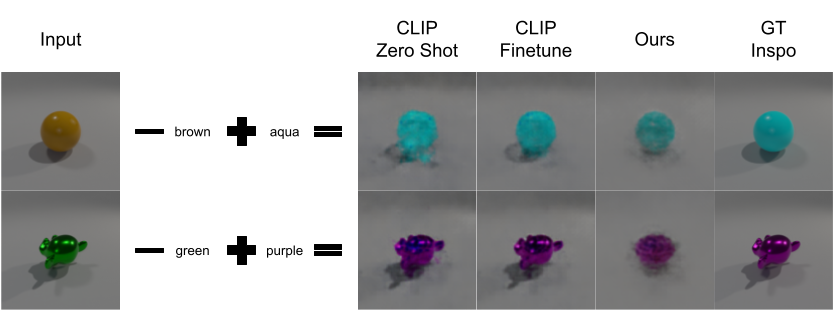}
    \caption{Color Editing}
  \end{subfigure}
  \hfill
  \begin{subfigure}[b]{0.9\textwidth}
    \includegraphics[width=\textwidth, center]{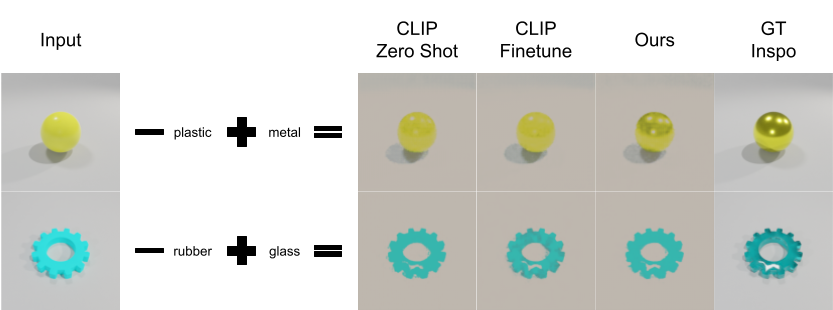}
    \caption{Material Editing}
  \end{subfigure}
  \hfill
  \begin{subfigure}[b]{0.9\textwidth}
    \includegraphics[width=\textwidth, center]{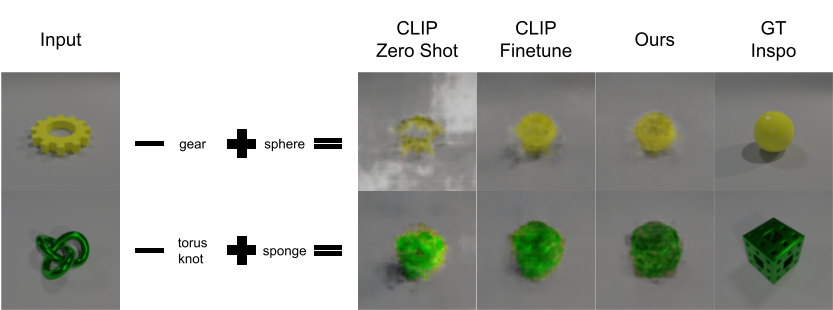}
    \caption{Shape Editing}
  \end{subfigure}
  \caption{More Qualitative Examples for Composition Reasoning}
  \label{appx_qr}
\end{figure*}

\end{document}